\documentclass[conference]{IEEEtran}

\usepackage{url}

\newcommand{\name}[1]{$\mathtt{#1}$}

\newtheorem{mymethod}{Method}
\newtheorem{mydefinition}{Definition}
\newtheorem{myassumption}{Assumption}
\newtheorem{myobservation}{Observation}
\newtheorem{myremark}{Remark}
\newtheorem{myproposition}{Proposition}
\newtheorem{myclaim}{Claim}
\newtheorem{mylemma}{Lemma}
\newtheorem{mycorollary}{Corollary}
\newtheorem{myexample}{Example}
\newtheorem{myexamples}{Examples}
\newtheorem{myalgorithm}{Algorithm}
\newtheorem{myconstruction}{Construction}
\newtheorem{myrule}{Rule}

\newcommand{\bolddot}{\hspace{-1.5mm}\textbf{.}\ \  }

\newcommand{\BT}{\begin{theorem}}
\newcommand{\ET}{\end{theorem}}
\newcommand{\BCR}{\begin{mycorollary}\bolddot}
\newcommand{\ECR}{\end{mycorollary}}
\newcommand{\BAS}{\begin{myassumption}}
\newcommand{\EAS}{\end{myassumption}}
\newcommand{\BPR}{\begin{myproposition}}
\newcommand{\EPR}{\end{myproposition}}
\newcommand{\BL}{\begin{mylemma}\bolddot}
\newcommand{\EL}{\end{mylemma}}
\newcommand{\BCM}{\begin{myclaim}\bolddot}
\newcommand{\ECM}{\end{myclaim}}

\newcommand{\BD}{\begin{mydefinition}}
\newcommand{\ED}{\end{mydefinition}}
\newcommand{\BPF}{\begin{proof}}
\newcommand{\EPF}{\end{proof}}
\newcommand{\BEX}{\begin{myexample}}
\newcommand{\EEX}{\end{myexample}}
\newcommand{\BEXS}{\begin{myexamples}}
\newcommand{\EEXS}{\end{myexamples}}
\newcommand{\BOB}{\begin{myobservation}}
\newcommand{\EOB}{\end{myobservation}}
\newcommand{\BR}{\begin{myremark}}
\newcommand{\ER}{\end{myremark}}
\newcommand{\BAL}{\begin{myalgorithm}}
\newcommand{\EAL}{\end{myalgorithm}}
\newcommand{\BAM}{\begin{mymethod}}
\newcommand{\EAM}{\end{mymethod}}

\newcommand{\BCO}{\begin{myconstruction}}
\newcommand{\ECO}{\end{myconstruction}}
\newcommand{\BRule}{\begin{myrule}}
\newcommand{\ERule}{\end{myrule}}

\newcommand{\BE}{\begin{enumerate}}
\newcommand{\EE}{\end{enumerate}}
\newcommand{\BI}{\begin{itemize}}
\newcommand{\EI}{\end{itemize}}

\newenvironment{Rightitem}{%
  \begin{itemize}}{\end{itemize}}
\newenvironment{Leftitem}{%
    \begin{itemize}}{\end{itemize}}
\newenvironment{Iffitem}{%
    \begin{itemize}}{\end{itemize}}
\newenvironment{Iffitemi}{%
    \begin{itemize}}{\end{itemize}}

\newcommand{\BRI}{\begin{Rightitem}\item}  \newcommand{\ERI}{\end{Rightitem}}
\newcommand{\BLI}{\begin{Leftitem}\item}   \newcommand{\ELI}{\end{Leftitem}}
\newcommand{\BIFF}{\begin{Iffitem}\item}   \newcommand{\EIFF}{\end{Iffitem}}
\newcommand{\BIFFI}{\begin{Iffitemi}\item}   \newcommand{\EIFFI}{\end{Iffitemi}}

\newcommand{\bc}{\begin{center}}
\newcommand {\ec}{\end{center}}

\newcommand{\stam}[1]{}

\newcommand{\bd}{\begin{definition}}
\newcommand{\ed}{\end{definition}}
\newcommand{\be}{\begin{enumerate}}
\newcommand{\bi}{\begin{itemize}}
\newcommand{\ee}{\end{enumerate}}
\newcommand{\ei}{\end{itemize}}

\usepackage{color, graphicx}

\begin{document}

\title{An Analysis of ISO 26262: Using Machine Learning Safely in Automotive Software}

\author{\IEEEauthorblockN{Rick Salay}
\IEEEauthorblockA{
University of Waterloo\\
ON, Canada\\
Email: rsalay@gsd.uwaterloo.ca
}
\and
\IEEEauthorblockN{Rodrigo Queiroz}
\IEEEauthorblockA{University of Waterloo\\
ON, Canada\\
Email:  rqueiroz@gsd.uwaterloo.ca}
\and
\IEEEauthorblockN{Krzysztof Czarnecki}
\IEEEauthorblockA{University of Waterloo\\
ON, Canada\\
Email:  kczarnec@gsd.uwaterloo.ca}}

\maketitle

\begin{abstract}
Machine learning (ML) plays an ever-increasing role in advanced automotive functionality for driver assistance and 
autonomous operation; however, its adequacy from the perspective of safety certification remains controversial. 
In this paper, 
we analyze the impacts that the use of ML as an implementation approach
has on ISO 26262 safety lifecycle and ask what could be done to address them. We
then provide a set of recommendations on how to adapt the standard to accommodate ML.
\end{abstract}

\IEEEpeerreviewmaketitle

\section{Introduction} \label{s:intro}
%

The use of machine learning (ML) is on the rise in many sectors of software development, and automotive software
development is no different. In particular, Advanced Driver Assistance Systems (ADAS) and Autonomous Vehicles (AV) 
are two areas where ML plays a significant role~\cite{spanfelner2012challenges, koopman2016challenges}.
In automotive development, safety is a critical objective, and the emergence of standards such as ISO 26262~\cite{ISO26262}
has helped focus industry practices to address safety in a systematic and consistent way. Unfortunately, 
ISO 26262 was not designed to accommodate technologies such as ML, and this has created a tension between
the need to innovate and the need to improve safety.

In response to this issue, research has been active in several areas. Recently, the safety of ML approaches in general have been analyzed both from theoretical~\cite{varshney2016engineering} and pragmatic perspectives~\cite{amodei2016concrete}. However, most 
research is specifically about neural networks (NN). Work on supporting the verification \& validation (V\&V) of NNs emerged in the 1990's with a focus on  making their internal structure easier to assess by extracting representations that are more understandable~\cite{tickle1998truth}. General V\&V methodologies for NNs
have also been proposed \cite{peterson1993foundation, rodvold1999software}.
More recently, with the popularity of deep neural networks (DNN), verification research has included more diverse topics such as
generating explanations of DNN predictions~\cite{hendricks2016generating}, improving the stability of classification~\cite{huang2016safety} and property checking of DNNs~\cite{katz2017reluplex}.  
%

Despite their challenges, NNs are already used in high assurance systems (see~\cite{schumann2010application} for a survey), 
and safety certification of NNs has received some attention. Pullum et al.~\cite{pullum2007guidance} give
detailed guidance on V\&V as well as other aspects of safety assessment such as \emph{hazard analysis} with a 
focus on adaptive systems in the the aerospace domain. Bedford et al.~\cite{bedford1996requirements} 
define general requirements for addressing NNs in any safety standard. Kurd et al.~\cite{kurd2007developing}
have established criteria for NNs to use in a safety case. 

The recent surge of interest in AV has also been driving research in certification. Koopman and Wagner~\cite{koopman2016challenges} identify some
of the key challenges to certification, including ML. Martin et al.~\cite{martin2017functional} analyze the adequacy of 
ISO 26262 for AV but focus on the impact of the increased complexity it creates rather than specifically the 
use of ML. Finally, Spanfelner et al.~\cite{spanfelner2012challenges} assess ISO 26262 from the perspective of 
driver assistance systems. 


The contribution of the current paper is complementary to the above research. We analyze the
impact that the use of ML-based software has on various parts of ISO 26262. Specifically,
we consider its impact in the areas of hazard analysis and in the phases of the software
development process.  In all, we identify five distinct problems that the use of ML creates
and make recommendations on steps toward addressing these problems both through
changes to the standard and through additional research.

The remainder of the paper is structured as follows. In Sec.~\ref{s:background}, we give the 
required background on ISO 26262 and ML. Sec.~\ref{s:anal} contains the analysis of the ISO 26262
safety lifecycle with five subsections describing each impacted area and the corresponding recommendations.
Finally, in Sec.~\ref{s:conc}, we summarize and give concluding remarks.

\section{Background} \label{s:background}

\subsection{ISO 26262}
\begin{figure} [h]
\centering{
\includegraphics[width=.5\textwidth]{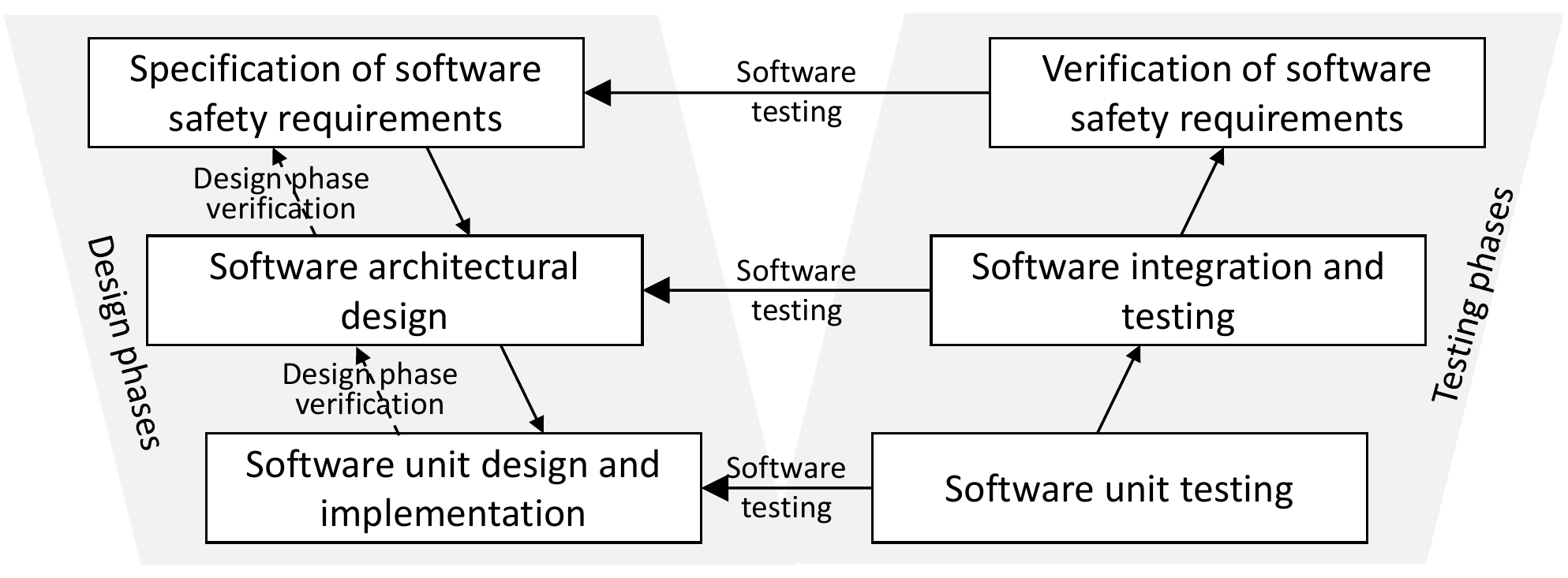}}
\vspace{-0.2in}
\caption{ISO 26262 part 6 - Product development at the software level.}
\label{f:vmodel}
\end{figure}

ISO 26262 is a standard that regulates functional safety of road vehicles. It recommends the use of a Hazard Analysis and Risk Assessment (HARA) method to identify hazardous events in the system and to specify safety goals that mitigate the hazards.  The standard has 10 parts, but we focus on Part 6: ``product development at the software level''. The standard follows the well-known \name{V} model for engineering shown in Fig.~\ref{f:vmodel}.

An Automotive Safety Integrity Level (ASIL) refers to a risk classification scheme defined in ISO 26262 for an item (e.g., subsystem) in an automotive system. The ASIL represents the degree of rigor required (e.g., testing techniques, types of documentation required, etc.) to reduce the risk of the item, where ASIL D represents the highest and ASIL A the lowest risk. If an element is assigned QM (Quality Management), it does not require  safety management.  The ASIL assessed for a given hazard is first assigned to the safety goal set to address the hazard and is then inherited by the safety requirements derived from that goal.  

\begin{figure} [h]
\centering{
\vspace{-0.1in}

\includegraphics[width=.5\textwidth]{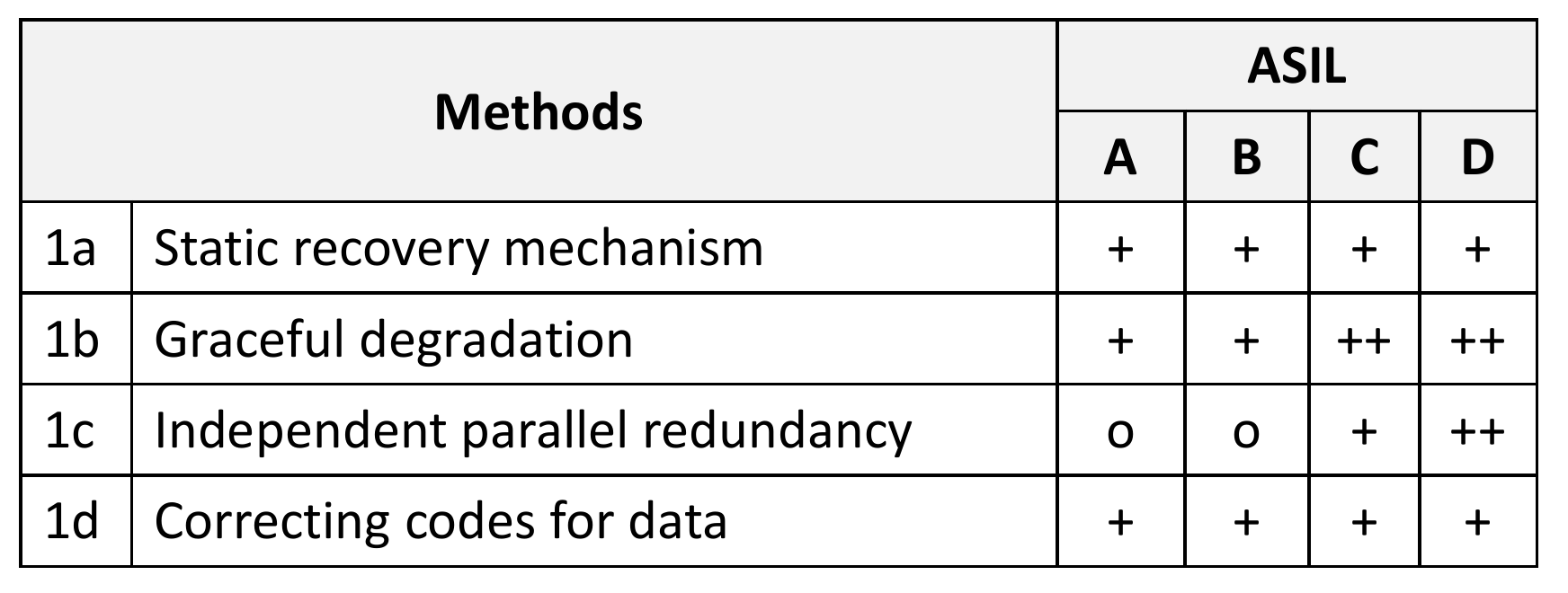}}
\caption{ISO 26262 Part 6 - Mechanisms for error handling at the software architectural level.}
\vspace{-0.1in}
\label{f:technique}
\end{figure}

Part 6 of the standard specifies the compliance requirements for software development.
For example, Fig.~\ref{f:technique} shows the error handling mechanisms recommended for use as part of the architectural design. The degree of recommendation for a method depends on the ASIL and is categorized as follows: “++” indicates that the method is highly recommended for the ASIL; “+” indicates that the method is recommended for the ASIL; and “o” indicates that the method has no recommendation for or against its usage for the ASIL. For example, \emph{Graceful Degradation} (1b) is the only highly recommended mechanism for an ASIL C item, while an ASIL D item would also require \emph{Independent Parallel Redundancy} (1c).

\subsection{Machine learning}
In this paper, we are concerned with software implementation using ML. We call a \emph{programmed component} to be one that is implemented using a programming language, regardless of whether the programming was done manually or automatically (e.g., via
code generation). In contrast, an \emph{ML component} is one that is a trained model using a supervised, unsupervised or 
reinforcement learning (RL) approach.  

There are several characteristics of ML that can impact safety or safety assessment.

{\bf Non-transparency.} All types of ML models contain knowledge in an encoded form, but this encoding is harder to interpret
for humans in some types than others. For example, a Bayesian Network for weather prediction is easier to interpret since the nodes are random variables representing human-defined concepts such as ``precipitation type'', ``temperature'', etc. In contrast, NN models are considered non-transparent, and significant research effort has been devoted to making them more transparent (e.g., \cite{tickle1998truth, hendricks2016generating}). Increasing ML model expressive power is typically at the expense of transparency but some research efforts focus on mitigating this~\cite{henzel2017}.  Non-transparency is an obstacle to 
safety assurance because it is more difficult for an assessor to develop confidence that the model is 
operating as intended.  

{\bf Error rate.} An ML model typically does not operate perfectly and exhibits some error rate. Thus, 
``correctness'' of an ML component, even with respect to test data, is seldom achieved and it must be assumed that it will 
periodically fail.
Furthermore, although an estimate of the true error rate is an output of the ML development process, there
is only a statistical guarantee about the reliability of this estimate. Finally, even if the estimate of the
true error rate was accurate, it may not reflect the error rate the system actually experiences while in operation after a
finite set of inputs because the true error is based on an infinite set of samples~\cite{varshney2016engineering}.
These characteristics must be considered when designing safe system using ML components.

{\bf Training-based.} Supervised and unsupervised learning based ML models are trained using a subset of possible inputs that  could be encountered operationally. Thus, the training set is necessarily incomplete and there is no guarantee that it is 
even representative of the space of possible inputs. In addition, learning may overfit a model by 
capturing details incidental to the training set rather than general to all inputs. RL suffers from similar limitations since it typically explores only a subset of possible behaviours during training. The uncertainty that this creates about
how an ML component will behave is a threat to safety. Another factor is that, even if the training set is representative, it may under-represent the safety-critical cases because these are often rarer in the input space~\cite{varshney2016engineering}.

{\bf Instability.} More powerful ML models (e.g., DNN) are typically trained using local optimization algorithms, and there can 
be multiple optima. Thus, even when the training set remains the same, the training process may
produce a different result. However, changing the training set also may change the optima. In general,
different optima may be far apart structurally, even if they are similar behaviourally. 
This characteristic makes it difficult to debug models or reuse parts of previous safety assessments.

\section{Analysis of ISO 26262}
\label{s:anal}



In this section, we detail our analysis of ML impacts on ISO 26262.
Since an ML component is a specialized type of software component, we define an area of the 
standard as \emph{impacted} when it is relevant to software components and the treatment of an ML component should differ 
from the existing treatment of software components by the standard. 
Applying this criterion to the ten parts of the standard resulted in identifying five areas of impact in two parts: 
the hazard analysis from the concept phase (Part 3) and the  
software development phase (Part 6). 
We describe the five areas of impact with corresponding recommendations in the following subsections. 

\subsection{Identifying hazards}
ISO 26262 defines a hazard as ``a potential source of harm caused by malfunctioning behaviour of the item where harm is physical injury or damage to the health of persons''~\cite[Part 1]{ISO26262}. 
The use of ML can create new types of hazards.
One type of such hazard is caused by the human operator becoming complacent because they think the automated driver assistance (often using ML) is smarter than it actually is~\cite{parasuraman1997humans}. For example, the driver stops monitoring steering in an automated steering function. On one level, this can be viewed as a case of ``reasonably forseeable misuse'' by the operator, and such misuse is identified in ISO 26262 as requiring mitigation~\cite[Part 3]{ISO26262}. 
However, this approach may be too simplistic. As ML creates opportunities for increasingly sophisticated driver assistance, the  role of the human operator becomes increasingly critical to correct for malfunctions. But increasing automation can create behavioural changes in the operator, reducing their skill level and limiting their ability to respond when needed~\cite{brookhuis2001behavioural}. Such behavioural impacts can negatively impact safety even though there is no system malfunction or misuse.  

Other new types of hazards are due to the unique ways an ML component can fail. For RL, faults in the reward
function can cause surprising failures. An RL-based component may negatively affect the environment in order to achieve its goal~\cite{amodei2016concrete}. For example, an AV may break laws in order to reach a destination
faster. Another possibility is that the RL component \emph{games} the reward function~\cite{amodei2016concrete}. For example, the AV figures out that it can avoid getting penalized for driving too close to other cars by exploiting certain sensor vulnerabilities so that it can't ``see'' how close it is getting. Although hazards such as these may be unique to ML components, they can be traced to faults, and thus they fit within the existing guidelines of ISO 26262.

{\bf Recommendations for ISO 26262}: 
The definition of hazard should be broadened to include harm potentially caused by complex behavioural interactions between humans and the vehicle that are not due to a system malfunction. 
The standard itself takes note that the current definition is ``restricted to the scope of ISO 26262; a more general definition is potential source of harm''\cite[Part 1]{ISO26262}. 
The definition and methods for identifying such hazards should be informed by the research specifically on
behavioural impacts of ADAS~\cite{sullivan2016literature} as well as human-robot interaction (HRI)\cite{goodrich2007human} more broadly.  For example, van den Brule et al.~\cite{van2014robot} study how a robot's behavioural style can affect the trust of humans interacting with it.

\subsection{Faults and failure modes}

ISO 26262 mandates the use of analyses such as Fault Mode Effects Analysis (FMEA) to identify how faults lead to failures that  may cause harm (i.e., are hazards). 
We can ask whether there are types of faults and failures that are unique to ML and not found in programmed software. 
Specific fault types and failure modes have been catalogued for NNs (e.g., ~\cite{pullum2007guidance, kurd2007developing}). 
Some of these are just ``apparent'' ML specific faults. For example,
a neuron that randomly changes its connection in an operational NN is not really about neurons but rather a conventional fault that can occur in the software on which the NN runs.
Others are distinctly ML-specific such as faults in the network topology, learning algorithm or training set. 
This creates the opportunity to develop focused tools and techniques to help find faults independently of the domain for 
which the ML model is being trained. 

Although ML faults have some unique characteristics, this cannot be said about failure modes.   
All faults can do is to increase the error rate of the deployed component, and thus cause one particular type of failure -- an incorrect output for some input. But since most software failures take the form of incorrect output for a given input, we may conclude that there is nothing different about the failure analysis of an ML component as compared to a programmed component, and 
existing ISO 26262 recommendations apply.

%
{\bf Recommendations for ISO 26262}: Require the use of fault detection tools and techniques that take into account the unique features of ML. For example, Chakarov et al.~\cite{chakarov2016debugging} describe a technique for debugging mis-classifications due to bad training in data, while Nushi et al.~\cite{nushi2016human} propose an approach for troubleshooting faults due
to complex interactions between linked ML components.

\subsection{The use of training sets}
Spanfelner et al.~\cite{spanfelner2012challenges} point out that there is an assumption in ISO 26262, given by the left side of the \name{V} model (Fig.~\ref{f:vmodel}), that component behaviour is fully specified and each refinement can be verified with respect to its specification. Note that this assumption is also made in other safety-critical domains such as aerospace~\cite{bhattacharyya2015certification}. 
This is important to ensure that a safety argument can trace the behaviour of the implementation to its design, safety requirements and ultimately, to the hazards that are mitigated.

This assumption is violated when a training set is used in place of a specification since such a set is necessarily incomplete, and it is not clear how to create assurance that the corresponding hazards are always mitigated.
Thus, an ML component violates the assumption.  
Furthermore, the training process is not a verification process since the trained model will be ``correct by construction'' with respect to the training set, up to the limits of the model and the learning algorithm.

A more careful analysis of the development lifecycle for an ML component shows that there are multiple levels
of specification and implementation, some of which may satisfy the assumption in the standard. High-level requirements for the component,
although abstract, can be expressed with completeness and traced to up to hazards. For example, the component may be required to 
``identify pedestrians'' that the AV should avoid harming. Detailed data requirements can be specified carefully to
ensure that an appropriate training, validation and testing sets are obtained. Subsequently, the data gathered can be verified with respect to this 
specification. Completeness is still an issue but \emph{coverage} can be used as a surrogate,
as it is with the design of test sets for software testing. 

A deeper issue, discussed by Spanfelner et al.~\cite{spanfelner2012challenges}, is that many kinds of advanced functionality require perception of the environment, and this functionality may be \emph{inherently unspecifiable}. For example, what
is the specification for recognizing a pedestrian? We might observe that since a vehicle 
must move around in a human world, advanced functionality must involve perception of \emph{human categories} (e.g., pedestrians). There is evidence that such categories can only partially be specified using rules (e.g., necessary and sufficient conditions) 
and also need examples~\cite{rouder2006comparing}. This suggests that ML-based approaches may actually be required
for implementing this type of functionality.

{\bf Recommendations for ISO 26262}:
The approach to safe implementation should be geared to the type of functionality being implemented. 
If the functionality is fully specifiable, then conventional programming can be required. In other
cases, such as advanced functionality requiring perception, ML-based approaches should be used,
and the complete specification requirement must be relaxed. 
Partial specifications can be required, where possible.
For example, if a pedestrian must be less than 9 feet tall, then
this property can be used to filter out false positives.
Such properties can be incorporated into the training process or checked on models
after training (e.g.,~\cite{katz2017reluplex}). 

Training set specifications and coverage metrics must be required to improve training set quality.
Ensemble methods such as boosting and decision fusion can also be recommended to improve the error rate. 
However, when the ASIL level is high, it is unlikely that the error rate can ever be brought to an acceptably low
level only through increasing or improving the
training set (due to the ``curse of dimensionality''). 
Therefore, fault tolerance strategies for software must be required.
For example, redundant pedestrian recognizers using different ML models and training sets can be used to detect
potential recognition failures when there is disagreement. 
Another possibility is to define a ``safety envelope'' of possible known safe behaviours and limit the ML component
to choose among them~\cite{perkins2002lyapunov}. 
Some of these recommendations may be addressed in a forth-coming OMG standard 
relating to sensor and perception issues~\cite{ISO21448}.

\subsection{Level of ML usage}
\label{s:level}
Fig.~\ref{f:vmodel} identifies an architectural level and a unit (i.e., component) level of implementation.
ISO 26262 defines a software architecture as consisting of components and their interactions in a 
hierarchical structure~\cite[Part 6]{ISO26262}. This component decomposition is important for safety because
it allows for easier comprehension of a complex system by human assessors and it permits the use of compositional
formal analysis techniques.

ML could be used to implement an entire software system, including its architecture, using an \emph{end-to-end} approach.
For example, Bojarski et al.~\cite{bojarski2016end} train a DNN to make the appropriate steering commands directly from raw sensor data, side-stepping typical AV architectural components such as lane detection, path planning, etc. 

Here, we may assume that the unit level, in the conventional sense of a distinct component that can be developed independently of the architecture, \emph{no longer exists}. This is the case, even if it is possible to extract and interpret the structure of the trained model as consisting of units with distinct functions, since this structure is emergent in the training process and unstable. If the model is re-trained with a slightly different training set, this structure can change arbitrarily. 
Note that a DNN does have an architecture in a different sense -- the set of layers and their connections. However, since
it is the training that actually ``implements'' the required functionality, this 
architecture is more of an generic execution layer.
Thus, an end-to-end approach deeply challenges the assumptions underlying ISO 26262. 

Another challenge with an end-to-end approach is that, in some cases, the the size of the training set needs to be exponentially larger than when a programmed architecture is used~\cite{shalev2016sample}. 
This puts additional strain on the already challenging problem of obtaining an adequate training set for safety-critical 
contexts. 

Finally, note that issues with an end-to-end approach can also apply when ML is used at the component level, if
components are too complex. For example, at one extreme, the architecture can consist of a single component. 
ISO 26262 specifically guards against this pitfall by mandating the use of modularity principles such as restricting
the size of components and maximizing the cohesion within a component. However, the lack of transparency of ML components can hamper the ability to assess component complexity and therefore, to apply these principles. Fortunately, 
improving ML transparency is an active research area (e.g.,~\cite{hendricks2016generating, tickle1998truth}).

{\bf Recommendations for ISO 26262}:
Although using an end-to-end approach has shown some recent successes with autonomous driving (e.g., ~\cite{bojarski2016end}),
we recommend that an end-to-end use of ML not be encouraged by ISO 26262, due to its incompatibility with the assumptions 
about stable hierarchical architectures of components.


\subsection{Required software techniques}
Part 6 of ISO 26262 deals with product development at the software level and specifies 75 software development techniques, such as shown in Fig.~\ref{f:technique}, that are used in various phases of the development process in the V model (Fig.~\ref{f:vmodel}).  Of these, 34 apply at the unit level, and the remaining at the architectural level.
We performed an assessment of the software techniques to determine their applicability to ML components\footnote{The 
data is available at \url{https://github.com/rsalay/safetyml}}.
Based on our recommendation in Sec~\ref{s:level}, we assumed that ML was only used at the unit level
and programming is used at the architecture level to connect components.

\begin{figure} [h]
\centering{
\includegraphics[width=.5\textwidth]{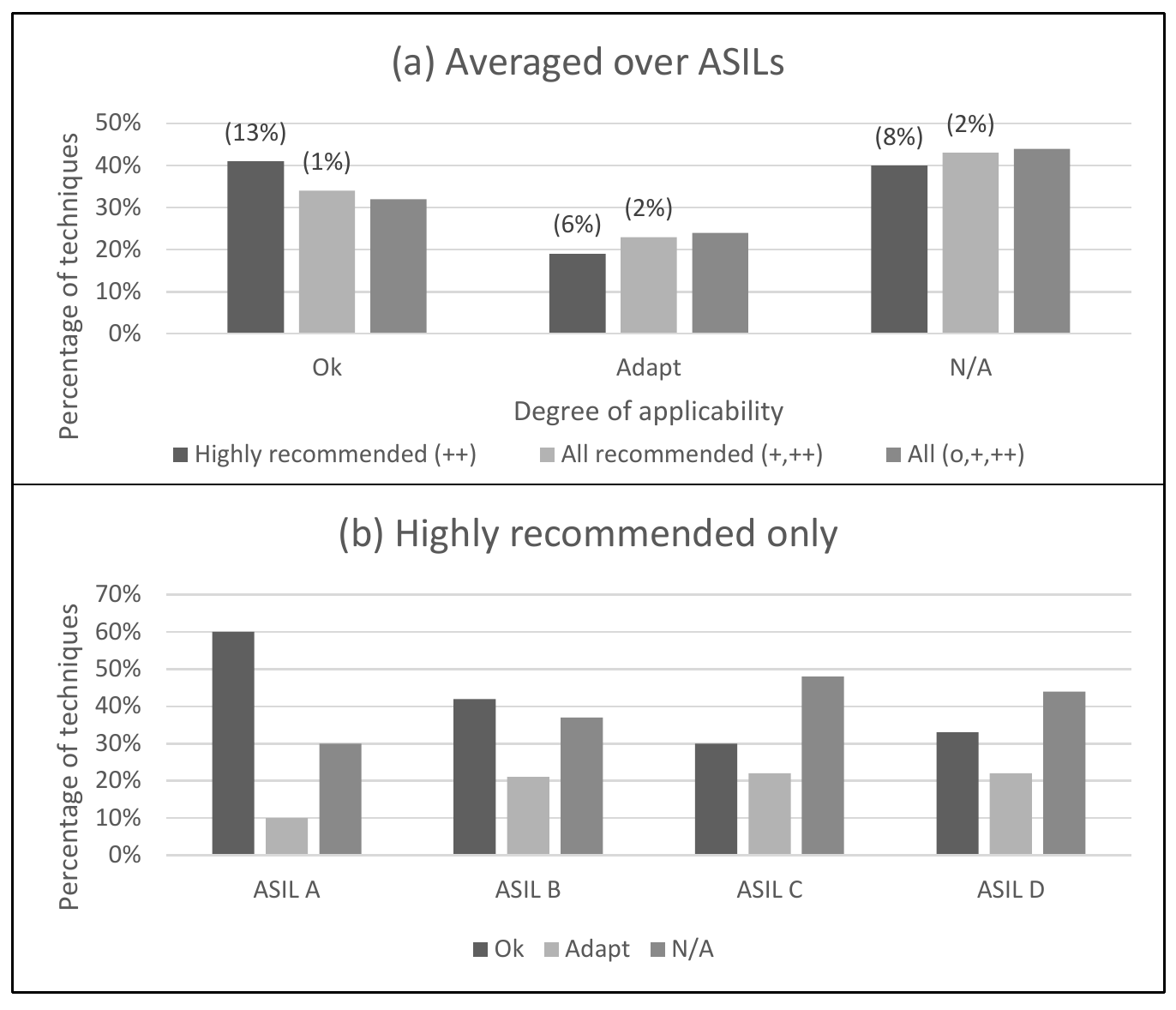}}
\vspace{-0.2in}
\caption{Percentage of unit-level software techniques applicable to ML components: (a) 
values averaged over the four ASIL levels with standard deviation shown in parentheses; (b) 
values for each ASIL when only highly recommended techniques are considered.
 }
\vspace{-0.2in}
\label{f:applic}
\end{figure}

The charts in Fig.~\ref{f:applic} show the results of the assessment for the techniques dealing with the unit level. 
We
classified each technique into one of three categories based on the level of applicability to ML. 
Category \name{Ok} means  
the technique is directly applicable without modification. Most of these cases are due to the fact
that they are black box techniques (e.g., \emph{analysis of boundary values}, \emph{error guessing}, etc.)
and thus, the method of component implementation is irrelevant. 
However, some white box techniques such as \emph{fault injection} also apply. For example, faults can be injected into an NN by breaking links or randomly changing weights~(e.g., \cite{takanami2000fault}). Category \name{Adapt} says that the technique can be used for an ML component
if it is adapted in some way. For example, the technique \emph{walk-through} can't be used
directly with an NN due to the non-transparency characteristic.
Finally, category \name{N/A} indicates that the technique is fundamentally code-oriented
and does not apply to an ML component. For example, \emph{no multiple use of variable names} is meaningful for 
a program but has no corresponding notion 
in an ML model. 

The results in Chart (a) are grouped by the degree to which the techniques are recommended. Recall from Sec.~\ref{s:background} that each technique is marked as highly recommended (++), recommended (+) or no recommendation (o) depending on the ASIL level. The bars in each category show the percentage of techniques that apply
when considering all techniques (0,+,++), only the recommended techniques (+, ++), and only the highly recommended techniques (++). 
Since the degree of recommendation varies by ASIL, each percentage is an average value over all four ASILs with the standard deviation in parentheses. Note that the standard deviation is 0 for the ``all'' group since every technique is present for each ASIL.
Because of the high standard deviation for the highly recommended group, we have included Chart (b) which gives the actual
data for each ASIL in this group.  

Chart (a) shows that a
significant part of the standard is still directly applicable (category \name{Ok}) 
and there is an emphasis on highly recommended techniques. However, the standard deviation is 
high and Chart (b) shows that most of these highly recommended techniques apply to the lower ASIL values -- i.e. they are less relevant from a safety critical perspective. Chart (a) also shows that about 40\% of the techniques do not apply
at all (category \name{N/A}) regardless of the degree of recommendation.
In general, techniques in the software part of the standard are clearly biased toward imperative 
programming languages (e.g., C, Java, etc.) \cite{bhattacharyya2015certification}. In addition to precluding
ML components, this bias makes it difficult to accept implementations in other mature programming
paradigms such as functional programming, logic programming, etc.  



{\bf Recommendations for ISO 26262}:
One approach to addressing the gap in applicable techniques as well as the imperative language bias without 
compromising safety may be to specify the requirements for techniques based on their \emph{intent}
and maturity rather than on their specific details. For example, the intent of the \emph{no multiple use of variable names} technique is to reduce the possibility for confusion that may prevent the detection
of bugs. This helps humans understand the implementation better and increase 
their confidence in its correctness and safety. Thus, the standard can require the use of ``accepted clarity increasing'' techniques instead of the specific techniques. 



\section{Summary and Conclusion} \label{s:conc}
%

Machine learning is increasingly seen as an effective software implementation technique for delivering advanced functionality;
however, how to assure safety when ML is used in safety critical systems is still an open question. 
The ISO 26262 standard for functional safety of road vehicles provides a comprehensive set of requirements for 
assuring safety but does not address the unique characteristics of ML-based software. In this paper, we
make a step towards addressing this gap by analyzing the places where ML can impact the standard and providing
recommendations on how to accommodate this impact. Our results and recommendations are summarized as follows.

{\bf Identifying hazards.} The use of ML can create new types of hazards that are not due to the malfunctioning of a component.
In particular, the complex behavioural interactions possible between humans and advanced functionality implemented
by ML can create hazardous situations that should be mitigated within the system design. We recommend that ISO 26262
expands their definition of hazard to address these kinds of situations.

{\bf Fault and failure modes.} ML components have a development lifecycle that is different from other types of software.
Analyzing the stages in the lifecycle reveals distinct types of faults they may have.
We recommend that ISO 26262 be extended to explicitly address the ML lifecycle and require the use of fault detection tools and techniques that are customized to this lifecycle.  

{\bf The use of training sets.} Because ML components are trained from inherently incomplete data sets, they violate the assumption in \name{V} model-based processes that 
component functionality must be fully specified and that refinements are verifiable. Furthermore, it is possible
that certain types of advanced functionality (e.g., requiring perception) for which ML is well suited are unspecifiable in principle. As a result, ML components are designed with the knowledge that
they have an error rate and that they will periodically fail. Rather than disqualifying this class of functionality, we 
recommend that ISO 26262 provide different safety requirements depending on whether the functionality is 
specifiable.

{\bf The level of ML usage.} ML could be used broadly at the architectural level with a system by using an end-to-end approach or
remain limited to use at the component level. The end-to-end approach challenges the assumption that a complex system is modeled
as a stable hierarchical decomposition of components each with their own function. This limits the use of most
techniques for system safety and we therefore recommend that ISO 26262 only allow the use of ML at the component level.

{\bf Required software techniques.} ISO 26262 mandates the use of many specific techniques for various stages of the software
development lifecycle. Our analysis shows that while some of these remain applicable to ML components and others
could readily be adapted, many remain that are specifically biased toward the assumption that code is implemented
using an imperative programming language. In order to remove this bias, we recommend that the requirements be 
expressed in terms of the intent and maturity of the techniques rather than their specific details.

\vspace{-0.05in}
\section*{Acknowledgment}
The authors would like to thank Atri Sarkar, Michael Smart, Michal Antkiewicz, Marsha Chechik, Sahar Kokaly and Ramy Shahin for their insightful comments.  

\vspace{-0.05in}
\bibliography{references}

\begin{thebibliography}{10}
\providecommand{\url}[1]{#1}
\csname url@samestyle\endcsname
\providecommand{\newblock}{\relax}
\providecommand{\bibinfo}[2]{#2}
\providecommand{\BIBentrySTDinterwordspacing}{\spaceskip=0pt\relax}
\providecommand{\BIBentryALTinterwordstretchfactor}{4}
\providecommand{\BIBentryALTinterwordspacing}{\spaceskip=\fontdimen2\font plus
\BIBentryALTinterwordstretchfactor\fontdimen3\font minus
  \fontdimen4\font\relax}
\providecommand{\BIBforeignlanguage}[2]{{%
\expandafter\ifx\csname l@#1\endcsname\relax
\typeout{** WARNING: IEEEtran.bst: No hyphenation pattern has been}%
\typeout{** loaded for the language `#1'. Using the pattern for}%
\typeout{** the default language instead.}%
\else
\language=\csname l@#1\endcsname
\fi
#2}}
\providecommand{\BIBdecl}{\relax}
\BIBdecl

\bibitem{spanfelner2012challenges}
B.~Spanfelner, D.~Richter, S.~Ebel, U.~Wilhelm, W.~Branz, and C.~Patz,
  ``{Challenges in applying the ISO 26262 for driver assistance systems},''
  \emph{Tagung Fahrerassistenz, M{\"u}nchen}, vol.~15, no.~16, p. 2012, 2012.

\bibitem{koopman2016challenges}
P.~Koopman and M.~Wagner, ``Challenges in autonomous vehicle testing and
  validation,'' \emph{SAE International Journal of Transportation Safety},
  vol.~4, no. 2016-01-0128, pp. 15--24, 2016.

\bibitem{ISO26262}
\emph{{ISO 26262: Road Vehicles -- Functional Safety}}, International
  Organization for Standardization, 2011, 1\textsuperscript{st} version.

\bibitem{varshney2016engineering}
K.~R. Varshney, ``Engineering safety in machine learning,'' \emph{arXiv
  preprint arXiv:1601.04126}, 2016.

\bibitem{amodei2016concrete}
D.~Amodei, C.~Olah, J.~Steinhardt, P.~Christiano, J.~Schulman, and D.~Man{\'e},
  ``{Concrete problems in AI safety},'' \emph{arXiv preprint arXiv:1606.06565},
  2016.

\bibitem{tickle1998truth}
A.~B. Tickle, R.~Andrews, M.~Golea, and J.~Diederich, ``The truth will come to
  light: Directions and challenges in extracting the knowledge embedded within
  trained artificial neural networks,'' \emph{IEEE Transactions on Neural
  Networks}, vol.~9, no.~6, pp. 1057--1068, 1998.

\bibitem{peterson1993foundation}
G.~E. Peterson, ``Foundation for neural network verification and validation,''
  in \emph{Optical Engineering and Photonics in Aerospace Sensing}.\hskip 1em
  plus 0.5em minus 0.4em\relax International Society for Optics and Photonics,
  1993, pp. 196--207.

\bibitem{rodvold1999software}
D.~M. Rodvold, ``A software development process model for artificial neural
  networks in critical applications,'' in \emph{Neural Networks, 1999.
  IJCNN'99. International Joint Conference on}, vol.~5.\hskip 1em plus 0.5em
  minus 0.4em\relax IEEE, 1999, pp. 3317--3322.

\bibitem{hendricks2016generating}
L.~A. Hendricks, Z.~Akata, M.~Rohrbach, J.~Donahue, B.~Schiele, and T.~Darrell,
  ``Generating visual explanations,'' in \emph{European Conference on Computer
  Vision}.\hskip 1em plus 0.5em minus 0.4em\relax Springer, 2016, pp. 3--19.

\bibitem{huang2016safety}
X.~Huang, M.~Kwiatkowska, S.~Wang, and M.~Wu, ``Safety verification of deep
  neural networks,'' \emph{arXiv preprint arXiv:1610.06940}, 2016.

\bibitem{katz2017reluplex}
G.~Katz, C.~Barrett, D.~Dill, K.~Julian, and M.~Kochenderfer, ``{Reluplex: An
  Efficient SMT Solver for Verifying Deep Neural Networks},'' \emph{arXiv
  preprint arXiv:1702.01135}, 2017.

\bibitem{schumann2010application}
J.~Schumann, P.~Gupta, and Y.~Liu, ``Application of neural networks in high
  assurance systems: A survey,'' in \emph{Applications of Neural Networks in
  High Assurance Systems}.\hskip 1em plus 0.5em minus 0.4em\relax Springer,
  2010, pp. 1--19.

\bibitem{pullum2007guidance}
L.~L. Pullum, B.~J. Taylor, and M.~A. Darrah, \emph{Guidance for the
  Verification and Validation of Neural Networks}.\hskip 1em plus 0.5em minus
  0.4em\relax John Wiley \& Sons, 2007, vol.~11.

\bibitem{bedford1996requirements}
D.~Bedford, G.~Morgan, and J.~Austin, ``Requirements for a standard certifying
  the use of artificial neural networks in safety critical applications,'' in
  \emph{Proceedings of the international conference on artificial neural
  networks}, 1996.

\bibitem{kurd2007developing}
Z.~Kurd, T.~Kelly, and J.~Austin, ``Developing artificial neural networks for
  safety critical systems,'' \emph{Neural Computing and Applications}, vol.~16,
  no.~1, pp. 11--19, 2007.

\bibitem{martin2017functional}
H.~Martin, K.~Tschabuschnig, O.~Bridal, and D.~Watzenig, ``{Functional Safety
  of Automated Driving Systems: Does ISO 26262 Meet the Challenges?}'' in
  \emph{Automated Driving}.\hskip 1em plus 0.5em minus 0.4em\relax Springer,
  2017, pp. 387--416.

\bibitem{henzel2017}
M.~Henzel, H.~Winner, and B.~Lattke, ``{Herausforderungen in der Absicherung
  von Fahrerassistenzsystemen bei der Benutzung maschinell gelernter und
  lernenden Algorithmen},'' in \emph{Proceedings of 11th Workshop
  Fahrerassistenzsysteme und automatisiertes Fahren (FAS)}, 2017, pp. 136--148.

\bibitem{parasuraman1997humans}
R.~Parasuraman and V.~Riley, ``Humans and automation: Use, misuse, disuse,
  abuse,'' \emph{Human Factors: The Journal of the Human Factors and Ergonomics
  Society}, vol.~39, no.~2, pp. 230--253, 1997.

\bibitem{brookhuis2001behavioural}
K.~A. Brookhuis, D.~De~Waard, and W.~H. Janssen, ``Behavioural impacts of
  advanced driver assistance systems--an overview,'' \emph{EJTIR}, vol.~1,
  no.~3, pp. 245--253, 2001.

\bibitem{sullivan2016literature}
J.~M. Sullivan, M.~J. Flannagan, A.~K. Pradhan, and S.~Bao, \emph{Literature
  Review of Behavioral Adaptations to Advanced Driver Assistance
  Systems}.\hskip 1em plus 0.5em minus 0.4em\relax AAA Foundation for Traffic
  Safety, 2016.

\bibitem{goodrich2007human}
M.~A. Goodrich and A.~C. Schultz, ``Human-robot interaction: a survey,''
  \emph{Foundations and Trends in Human-Computer Interaction}, vol.~1, no.~3,
  pp. 203--275, 2007.

\bibitem{van2014robot}
R.~van~den Brule, R.~Dotsch, G.~Bijlstra, D.~H. Wigboldus, and P.~Haselager,
  ``Do robot performance and behavioral style affect human trust?''
  \emph{International Journal of Social Robotics}, vol.~6, no.~4, pp. 519--531,
  2014.

\bibitem{chakarov2016debugging}
A.~Chakarov, A.~Nori, S.~Rajamani, S.~Sen, and D.~Vijaykeerthy, ``Debugging
  machine learning tasks,'' \emph{arXiv preprint arXiv:1603.07292}, 2016.

\bibitem{nushi2016human}
B.~Nushi, E.~Kamar, E.~Horvitz, and D.~Kossmann, ``{On Human Intellect and
  Machine Failures: Troubleshooting Integrative Machine Learning Systems},''
  \emph{arXiv preprint arXiv:1611.08309}, 2016.

\bibitem{bhattacharyya2015certification}
S.~Bhattacharyya, D.~Cofer, D.~Musliner, J.~Mueller, and E.~Engstrom,
  ``Certification considerations for adaptive systems,'' in \emph{Unmanned
  Aircraft Systems (ICUAS), 2015 International Conference on}.\hskip 1em plus
  0.5em minus 0.4em\relax IEEE, 2015, pp. 270--279.

\bibitem{rouder2006comparing}
J.~N. Rouder and R.~Ratcliff, ``{Comparing exemplar and rule-based theories of
  categorization},'' \emph{Current Directions in Psychological Science},
  vol.~15, no.~1, pp. 9--13, 2006.

\bibitem{perkins2002lyapunov}
T.~J. Perkins and A.~G. Barto, ``{Lyapunov design for safe reinforcement
  learning},'' \emph{Journal of Machine Learning Research}, vol.~3, no. Dec,
  pp. 803--832, 2002.

\bibitem{ISO21448}
\emph{{ISO/AWI PAS 21448: Road Vehicles -- Safety of the Intended
  Functionality}}, International Organization for Standardization, (under
  development).

\bibitem{bojarski2016end}
M.~Bojarski, D.~Del~Testa, D.~Dworakowski, B.~Firner, B.~Flepp, P.~Goyal, L.~D.
  Jackel, M.~Monfort, U.~Muller, J.~Zhang \emph{et~al.}, ``End to end learning
  for self-driving cars,'' \emph{arXiv preprint arXiv:1604.07316}, 2016.

\bibitem{shalev2016sample}
S.~Shalev-Shwartz and A.~Shashua, ``On the sample complexity of end-to-end
  training vs. semantic abstraction training,'' \emph{arXiv preprint
  arXiv:1604.06915}, 2016.

\bibitem{takanami2000fault}
I.~Takanami, M.~Sato, and Y.~P. Yang, ``{A fault-value injection approach for
  multiple-weight-fault tolerance of MNNs},'' in \emph{Proceedings of the
  IEEE-INNS-ENNS International Joint Conference on Neural Networks},
  vol.~3.\hskip 1em plus 0.5em minus 0.4em\relax IEEE, 2000, pp. 515--520.

\end{thebibliography}



\bibliographystyle{IEEEtran}
%

\end{document}